\documentclass[conference]{IEEEtran}
\IEEEoverridecommandlockouts
\usepackage{cite}
\usepackage{amsmath,amssymb,amsfonts}
\usepackage{algorithmic}
\usepackage{graphicx}
\usepackage{textcomp}
\usepackage{xcolor}
\usepackage[]{lipsum}
\usepackage{graphicx}
\usepackage{threeparttable}
\usepackage{booktabs}
\usepackage{multirow}
\usepackage{caption}
\usepackage[ruled,vlined]{algorithm2e}
\usepackage{array}
\usepackage{dblfloatfix}
\usepackage{float}
\usepackage{cite}
\usepackage[normalem]{ulem}
\usepackage[caption=false,font=normalsize,labelfont=sf,textfont=sf]{subfig}

\usepackage{cleveref}
\crefname{section}{§}{§§}
\Crefname{section}{§}{§§}

\usepackage{xcolor}  
\usepackage{soul}  

\def\BibTeX{{\rm B\kern-.05em{\sc i\kern-.025em b}\kern-.08em
    T\kern-.1667em\lower.7ex\hbox{E}\kern-.125emX}}
\begin{document}

\title{Federated Variational Learning for Anomaly Detection in Multivariate Time Series\\
%{\footnotesize \textsuperscript{*}Paper Review Version in AFRL}
\thanks{*These authors contributed equally to this work}
\thanks{978-1-6654-4331-9/21/\$31.00 ©2021 IEEE}}

\author{%
    Kai Zhang$^{1*}$, Yushan Jiang$^{1*}$, Lee Seversky$^{2}$, Chengtao Xu$^{1}$, Dahai Liu$^{1}$, Houbing Song$^{1}$\\
    $^{1}$Embry-Riddle Aeronautical University, Daytona Beach, FL 32114\\
    $^{2}$Air Force Research Laboratory, Rome, NY 13441\\
    $^{1}$\{zhangk3, jiangy2, xuc3\}@my.erau.edu, dahai.liu@erau.edu, h.song@ieee.org\\
    $^{2}$lee.seversky@us.af.mil
}

\maketitle
\makeatletter
\def\ps@IEEEtitlepagestyle{%
  \def\@oddfoot{\mycopyrightnotice}%
  \def\@oddhead{\hbox{}\@IEEEheaderstyle\leftmark\hfil\thepage}\relax
  \def\@evenhead{\@IEEEheaderstyle\thepage\hfil\leftmark\hbox{}}\relax
  \def\@evenfoot{}%
}
\def\mycopyrightnotice{%
  \begin{minipage}{\textwidth}
  \centering \scriptsize
  Copyright~\copyright~2021 IEEE. Personal use of this material is permitted. Permission from IEEE must be obtained for all other uses, in any current or future media, including\\reprinting/republishing this material for advertising or promotional purposes, creating new collective works, for resale or redistribution to servers or lists, or reuse of any copyrighted component of this work in other works by sending a request to pubs-permissions@ieee.org.
  \end{minipage}
}
\makeatother
 
\begin{abstract}
Anomaly detection has been a challenging task given high-dimensional multivariate time series data generated by networked sensors and actuators in Cyber-Physical Systems (CPS). Besides the highly nonlinear, complex, and dynamic nature of such time series, the lack of labeled data impedes data exploitation in a supervised manner and thus prevents an accurate detection of abnormal phenomenons. On the other hand, the collected data at the edge of the network is often privacy sensitive and large in quantity, which may hinder the centralized training at the main server. To tackle these issues, we propose an unsupervised time series anomaly detection framework in a federated fashion to continuously monitor the behaviors of interconnected devices within a network and alert for abnormal incidents so that countermeasures can be taken before undesired consequences occur. To be specific, we leave the training data distributed at the edge to learn a shared Variational Autoencoder (VAE) based on Convolutional Gated Recurrent Unit (ConvGRU) model, which jointly captures feature and temporal dependencies in the multivariate time series data for representation learning and downstream anomaly detection tasks. Experiments on three real-world networked sensor datasets illustrate the advantage of our approach over other state-of-the-art models. We also conduct extensive experiments to demonstrate the effectiveness of our detection framework under non-federated and federated settings in terms of overall performance and detection latency.

% \hl{import matplotlib
% matplotlib.rcParams['pdf.fonttype'] = 42
% matplotlib.rcParams['ps.fonttype'] = 42}

\end{abstract}


\section{Introduction}
Anomaly detection has been an important research topic in data mining community, where fruitful progress has been made with a wide range of critical applications in emergent Cyber-Physical Systems (CPS). Efficient and accurate anomaly detection helps to monitor system status continuously and alert for potential incidents so that interposition can be timely executed to mitigate the negative consequential impact \cite{ren2019time}. Meanwhile, a massive amount of multivariate time series data become available with the rising of real-world sensor embedded systems, which favors the utilization of data-driven detection methods. However, this task is challenging due to the nature of collected data from inter-connected sensor networks: the collected data is usually highly imbalanced with very few labeled anomalies \cite{lin2020anomaly}, hindering the usage of supervised learning methods which require sufficient labeled data for model training \cite{su2019robust}; furthermore, it is hard to detect unseen anomalies without a known priori. Due to these constraints, unsupervised learning appears to be the desired choice in our methodology to detect anomalous events.  

%\noindent \hl{variable $\rightarrow$ metrics?}

A rich body of literature exists on detecting multivariate time series anomalies in an unsupervised way (see \cref{sec:related_work}). The core idea behind these approaches is to \textit{learn robust latent representations to capture normal patterns of multivariate time series considering various dependencies.} The more different a sample is from the normal patterns, the more likely it is considered as an anomaly \cite{su2019robust}. Although many effective algorithms especially deep learning-based methods have been proposed, the authors did not consider the aspect that heterogeneous behaviors of various networks in CPS exist in the form of isolated islands for privacy reasons. On the other hand, newly-deployed devices cannot provide sufficient data in a short time for training a robust anomaly detection model by themselves. 

To deal with the data scarcity and privacy-preserving problems, we propose \textit{FedAnomaly}, an unsupervised federated anomaly detection framework based on Variational Auto-Encoder (VAE) with elaborated Convolutional GRU (ConvGRU). Our approach innovatively glues federated learning and deep generative model with the following two key techniques: (a) %Shi et al. \cite{shi2015convolutional} has shown that having convolutional structures in both the input-to-state and state-to-state transitions can capture more spatiotemporal correlations. Inspired by it, 
We propose to use ConvGRU, a convolution gated sequential learning model, to explicitly and jointly capture temporal and feature-level dependencies in the multivariate time series for representation learning and downstream anomaly detection tasks; and (b) To help capture robust representations of time series at cross-silos, we adopted VAE \cite{kingma2013auto} with Federated Learning (FL) \cite{mcmahan2017communication}, a paradigm aiming at building machine learning models based on distributed datasets across multiple devices while preventing data leakage \cite{yang2019federated}, into an anomaly detection system. In this system, VAE offers a combination of highly flexible non-linear mapping between the latent random state and the observed output that results in effective approximate inference \cite{chung2015recurrent}.
%https://stats.stackexchange.com/questions/324340/when-should-i-use-a-variational-autoencoder-as-opposed-to-an-autoencoder
The main contributions of this paper are:

\begin{itemize}
    \item We employ a federated deep generative model for time series anomaly detection for the first time. 
    \item This is the first work leveraging ConvGRU in an outlier detection system to jointly capture the underlying feature-level and temporal patterns of multivariate time series. The experiments on three real-world datasets show that our approach outperforms the state-of-the-art model regarding F1 scores.
    \item Further experiments and analysis demonstrate the effectiveness of our detection framework under non-federated and federated settings in terms of overall evaluation metrics, the detection of contiguous anomaly segments, and detection latency.
\end{itemize}

The remainder of this paper is organized as follows: Section \ref{sec:related_work} reviews cutting-edge methods for multivariate time series anomaly detection. Section \ref{sec:method} presents our framework. The experimental setup and results are described in Section \ref{sec:experiments}, demonstrating the effectiveness of our framework in multiple aspects. Section \ref{sec:conclusion} concludes our paper.

\section{Related Work} \label{sec:related_work}
Abundant literature focuses on multivariate time series anomaly detection. Broadly, they can be classified into two categories, namely \textit{predictive models} and \textit{generative models}. In this section, we first review the state-of-the-art approaches based on these two paradigms to detect anomalies in multivariate time series, then discuss pioneering work about outlier identification using federated learning architecture.  

\subsection{Predictive Models}
A predictive model detects anomalies in terms of prediction error \cite{chen2021learning, deng2021graph, lin2020anomaly, hundman2018detecting, ren2019time}. Variations of recurrent neural networks (RNNs) such as Long Short-Term Memory (LSTM) \cite{hochreiter1997long} and Gated Recurrent Unit (GRU) \cite{chung2014gru} are the most popular methods to predict the next single or multiple time steps in anomaly detection. For example, \cite{hundman2018detecting} applied LSTM for multivariate time series prediction and identified anomalies in the spacecraft using prediction error with dynamic thresholds. Different from forecasting the future trend of time series, Lin et al. \cite{lin2020anomaly} extracted the embedding of the local information over a short window using VAE and employed LSTM to predict the next window's embedding. %Although they argued that their hierarchical structure contributes to detect anomalies occurring over both short and long periods, the simple comparison with baselines cannot support their findings. 
Capturing temporal patterns is not the only way to identify outliers in a sequence. Microsoft \cite{ren2019time} combined spectral residual (SR) and convolutional neural network (CNN) for time series anomaly detection, in which SR captures the spectrum information (a saliency map) of a sequence, and CNN classifies anomalies. Recently, graph neural network for anomaly detection has been an active research topic since it is intractable for conventional machine learning methods to model complicated dependencies among entities in a network. Deng et al. \cite{deng2021graph} used a graph structured learning approach with graph attention network to predict the expected behavior of time series, additionally providing explainability for the detected anomalies based on attention weights. Chen et al. \cite{chen2021learning} proposed the influence propagation convolution to model the anomaly information flow between graph nodes for multivariate time series anomaly detection.

\subsection{Generative Models}
The core idea of generative models is to learn the representation of normal patterns instead of anomalies in a time series. The state-of-the-art deep generative models for anomaly detection include DAGMM \cite{zong2018deep}, VAE \cite{xu2018unsupervised, su2019robust}, and GAN \cite{li2019mad}. Generally speaking, they all first recognize normal regions in some latent feature spaces, and then evaluate how far an observation is from the normal regions to detect anomalies. \cite{xu2018unsupervised}. Due to the mode collapse and non-convergence problem of GAN, \cite{audibert2020usad} adopted an adversarial strategy to amplify the reconstruction error of inputs containing anomalies. To be specific, two autoencoders are trained to reconstruct normal input, then they are trained in an adversarial way where one autoencoder will seek to fool the other one. A few researchers endeavored to integrate the advantages of predictive and generative manners \cite{zhao2020multi, lin2020anomaly}. For instance, \cite{zhao2020multi} leveraged a joint optimization target in terms of prediction error and reconstruction error, specifically, the predictive component based on GRU forecasts the value at the next step while the generative component based on VAE obtains the distribution of entire time series. 

It is noteworthy that DAGMM \cite{zong2018deep}, which was not designed for multivariate time series, neglects the inherent temporal dependencies of time series. On the contrary, prior works on VAE \cite{xu2018unsupervised, su2019robust} and GAN \cite{li2019mad} only consider temporal dependency while not explicitly addressing latent interactions among features. 
Moreover, despite the great promise of generative models in anomaly detection, simply replacing the feed-forward layer with RNN in a VAE does not perform well according to our experimental results (see \cref{sec:comparison}). Along this direction, we are interested in developing a generative model which can capture both feature-level and temporal dependencies (see \cref{sec:conv1dgru}). 

\subsection{Federated Learning for Anomaly Detection} \label{sec:fl_anomaly}
Federated learning (FL) enables a large amount of edge computing devices to jointly learn a model without data sharing \cite{li2019convergence}. As a well-known algorithm, FedAvg \cite{mcmahan2017communication} runs stochastic gradient descent (SGD) on local devices (say on randomly selected clients) and averages updates per round of communication between parameter server and clients. Such a setting empowers training a robust model for detecting anomalies in CPS since it tackles the data scarcity problem from a privacy-preserving perspective. D{\"I}oT \cite{nguyen2019diot} firstly employed FL approach to anomaly-detection-based intrusion detection. To alleviate the heterogeneity of network traffic packets from different devices, they mapped packet sequences as symbols, which are then fed into pre-trained GRU model to determine possible intrusions by estimating the occurrence probabilities of each symbol based on its preceding symbols. Communication is the other primary bottleneck for FL since edge devices typically operate at lower rate links that are potentially expensive and unreliable \cite{mcmahan2021advances}. Therefore, \cite{liu2020deep} applied sparsification technique %\cite{wangni2017gradient} 
to achieve compressed gradients for reducing communication cost. Moreover, \cite{zhao2019multi} proposed multi-task federated learning for network anomaly detection, which helps provide more network information such as traffic characteristics to human operators. Different from aforementioned works that focus on federated predictive models, \cite{chen2019network} adopted a generative model, DAGMM, in a federated fashion. We also propose a federated generative model in the paper, but we use ConvGRU-VAE to capture feature and temporal dependencies jointly. Moreover, we evaluate our model with real-world and high-dimensional network datasets instead of redundant and non-sequential data (say KDD-99) %\cite{tavallaee2009detailed} 
like \cite{chen2019network}.

\section{Methodology} \label{sec:method}
In this section, we present the problem statement of multivariate time series anomaly detection in the federated learning setting, followed by the overall structure of our system. In addition, we provide an introduction to the key components of our framework in detail.

\begin{figure}
    \centering
    \includegraphics[height=4cm]{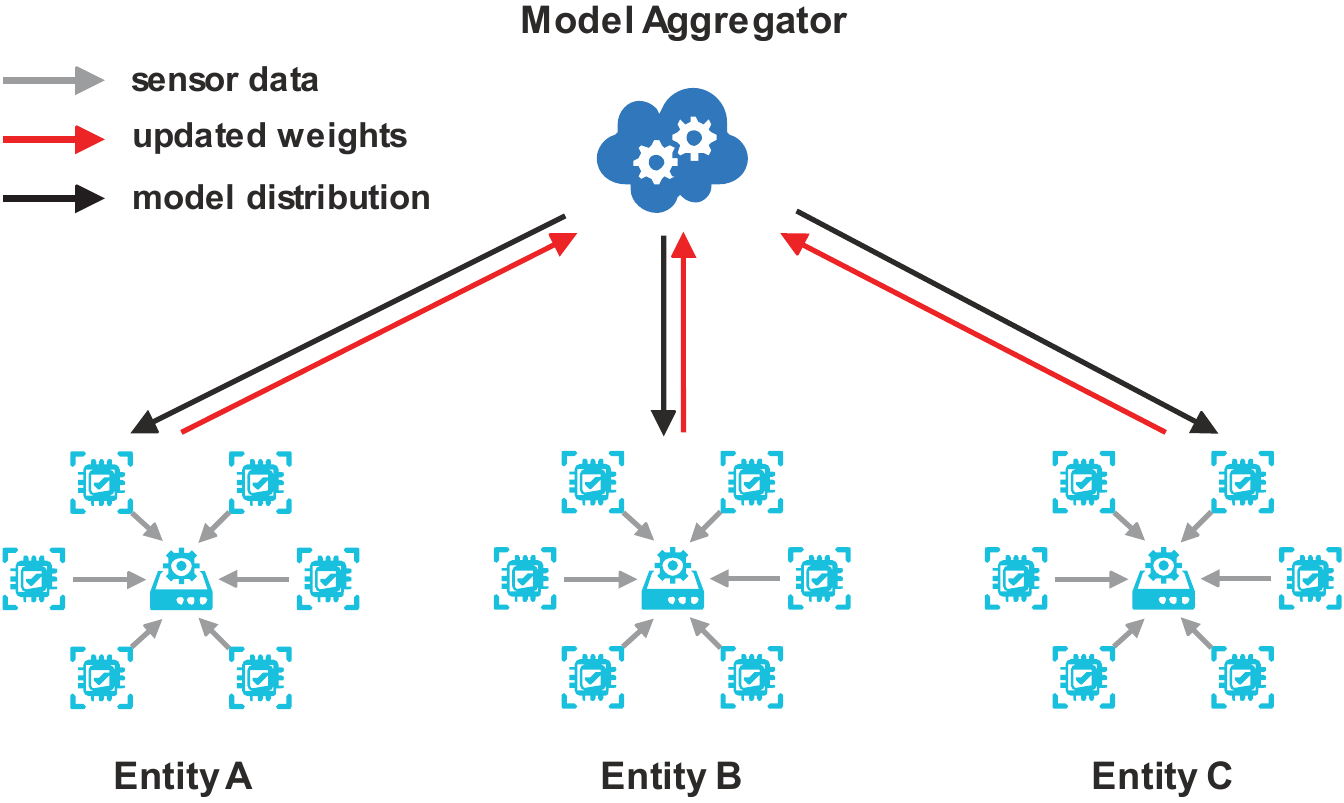}
    \caption{\small{The overview of the federated learning paradigm: In our problem setting, each sensor or edge device measures the metric and sends it to the entity server (client). Afterward, the entity server trains the local model on its collected network metrics, and uploads the gradients to the cloud aggregator. There are two procedures on the aggregator side: (a) it initializes the global model and distributes it to edges, and (b) it aggregates the uploaded parameters and sends them back to clients.}}
    \label{fig:overview}
    \vspace{-0.5cm}
\end{figure}

\subsection{Problem Statement}

In this paper, our objective is to find anomalies in entities based on multivariate time series anomaly detection, which is divided into two stages -- training stage and detection stage. 

In the training stage, our goal is to learn the representation of normal time series. As shown in Fig. \ref{fig:overview}, the training set consists of the sensor and/or actuator data over a certain period from different entities. To be specific, the training data is defined as: ${\boldsymbol{X}_{train}}=\left(\boldsymbol{X}_{1}, \boldsymbol{X}_{2}, \ldots, \boldsymbol{X}_{L}\right)^{T} \in \mathbb{R}^{ N \times L \times F}$, where $N$ denotes the number of entities, $F$ denotes the number of (sensor) metrics of each entity, and $L$ denotes the number of timestamps in a training sequence. For each timestamp $t$, $\boldsymbol{X}_{t}=\left(\boldsymbol{x}_{t}^{1}, \boldsymbol{x}_{t}^{2}, \ldots, \boldsymbol{x}_{t}^{N}\right)^{T} \in \mathbb{R}^{ N \times F}$ denotes the status of the whole network with all metrics of all entities, where $\boldsymbol{x}_{t}^{i}=\left(\boldsymbol{x}_{t}^{i,1}, \ldots, \boldsymbol{x}_{t}^{i,F}\right)^{T} \in \mathbb{R}^{F}$ denotes all metrics of entity $i$ at timestamp $t$ and $\boldsymbol{x}_{t}^{i,j} \in \mathbb{R}$ denotes the value of the corresponding $j$-th metric. Considering the commonly used formulation in anomaly detection tasks and the existence of high imbalance between normal and anomalous samples in the most realistic scenarios, the training data on the client is set to contain normal data only without any anomalies. 

In the detection stage, our target is to find anomalies in the testing data which is from the same entities with the same metrics but collected from a different period. Similar to the training data, the testing data is defined as ${\boldsymbol{X}_{test}}=\left(\boldsymbol{X}_{1}, \boldsymbol{X}_{2}, \ldots, \boldsymbol{X}_{L'}\right)^{T} \in \mathbb{R}^{ N \times {L'} \times F}$, where $L'$ is the length of a testing sequence. The testing data is usually distributed on different entities in practice. For multivariate time series anomaly detection, at a timestamp $t'$, the preceding sequential observations ${\boldsymbol{X}_{t'-L':t'}}$ are used to determine whether an observation ${\boldsymbol{X}_{t'}}$ is anomalous ($y_{t'} = 1$) or not ($y_{t'} = 0$).

%the output of our detection framework is a set of binary labels indicating the corresponding status with anomaly or not, denoted as $\boldsymbol{Y}_{t'} \in {(0,1)}$, where $y_{t'} = 1$ indicates an abnormal status. The whole output in the testing set is then denoted as $\boldsymbol{Y}_{test}=\left(\boldsymbol{Y}_{1}, \boldsymbol{Y}_{2}, \ldots, \boldsymbol{Y}_{L'}\right)^{T} \in \mathbb{R}^{ {L'} \times 1}$.

\subsection{Overall Structure of FedAnomaly} \label{sec:fedanomaly}
\begin{figure}
    \centering
    \includegraphics[height=4cm]{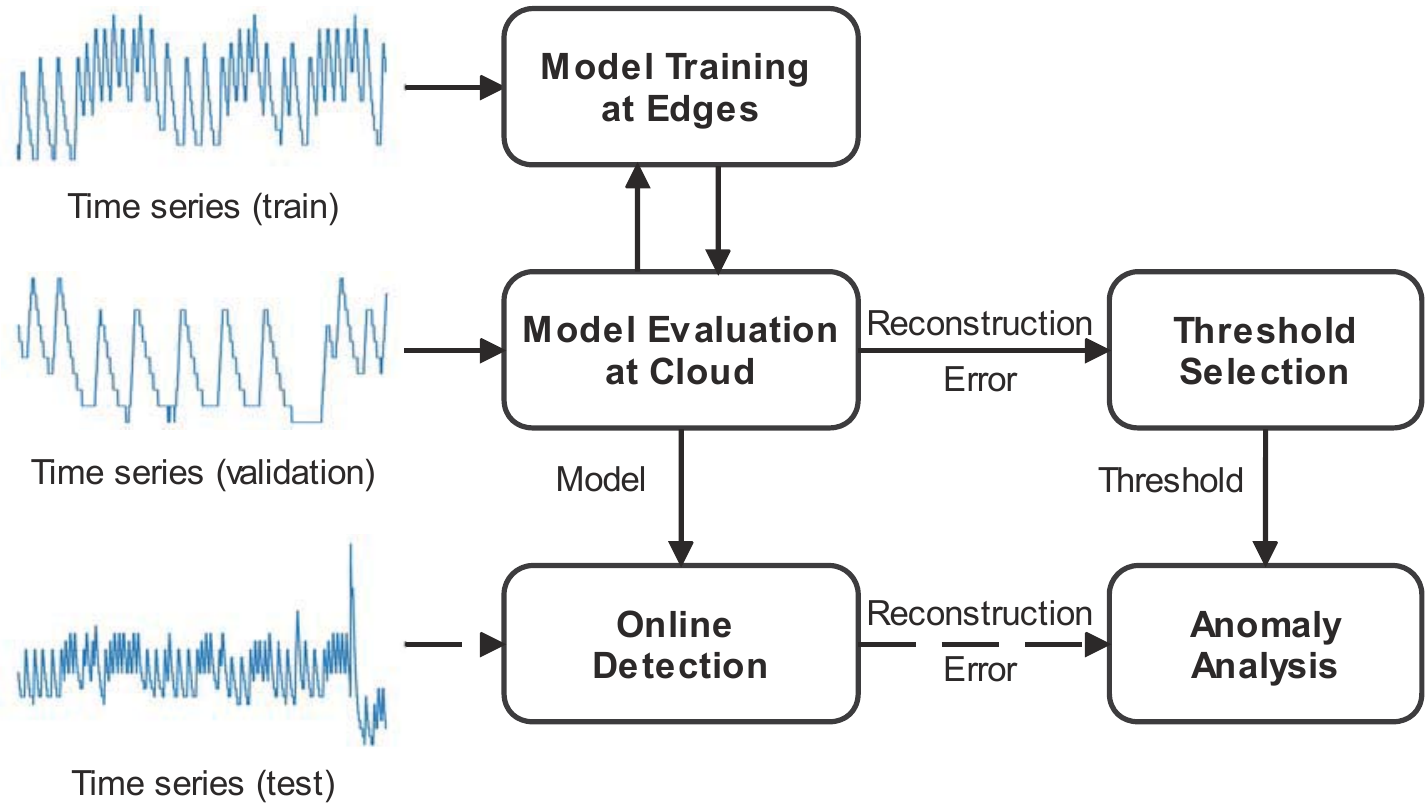}
    \caption{\small{The overall structure of FedAnomaly: For simplicity, we use a univariate time series to represent a multivariate input in the figure. The solid lines denote collaborative training and the dash lines show the pipeline of online detection.}}
    \label{fig:structure}
    \vspace{-0.5cm}
\end{figure}

Fig. \ref{fig:structure} shows the overall structure of FedAnomaly consisting of two parts: collaborative training and online detection. Note that the data preprocessing module, which transforms, standardizes and then segments data into sequences through a sliding window of a fixed length, is omitted in the figure. After preprocessing, multivariate time series are fed into local models to capture the patterns of training data. Model aggregator updates parameters of the global model by aggregating gradients from edges, and outputs a reconstruction error of an observation at the last timestamp in a sequence. A standard or validation data is stored on the cloud, and we continuously train the global model until the reconstruction error of the standard data converges. These reconstruction errors are utilized by the threshold selection module to choose an anomaly threshold for online detection. To be specific, we select the maximum reconstruction error in the standard data as the threshold. The online detection module on each edge receives the trained model and threshold from the cloud. Entities now can obtain the anomaly result of a new observation. If the reconstruction error of ${\boldsymbol{X}_{t'}}$ is below the anomaly threshold, ${\boldsymbol{X}_{t'}}$ will be declared as anomalous, otherwise, it is normal.

\subsection{Convolutional Gated Recurrent Unit (ConvGRU)} \label{sec:conv1dgru}

For general sequence learning tasks, LSTM and GRU have shown the advantages in terms of stability and capability of modeling long-range temporal dependencies. Meanwhile, there are more and more variants that have been designed to model more complicated correlations for downstream tasks. Enlightened by \cite{shi2015convolutional,siam2017convolutional} and their success on learning the representation of a spatial-temporal sequence for weather forecasting and video segmentation tasks, we also leverage the kernel-gated RNNs to capture the dependencies within multivariate time series as the representation learning is ongoing. To be specific, we apply a convolutional kernel in input-to-state and state-to-state transitions on the basic Gated Recurrent Unit, and implement a variant of this sequence learning model, Convolutional Gated Recurrent Unit (ConvGRU) so that the underlying feature-level dependencies can be captured along with temporal dependencies. As such, a better representation of multivariate time series can be generated under the encoder-decoder framework for the anomaly detection task. 

\begin{figure}
  \centering
  \includegraphics[height=4.5cm]{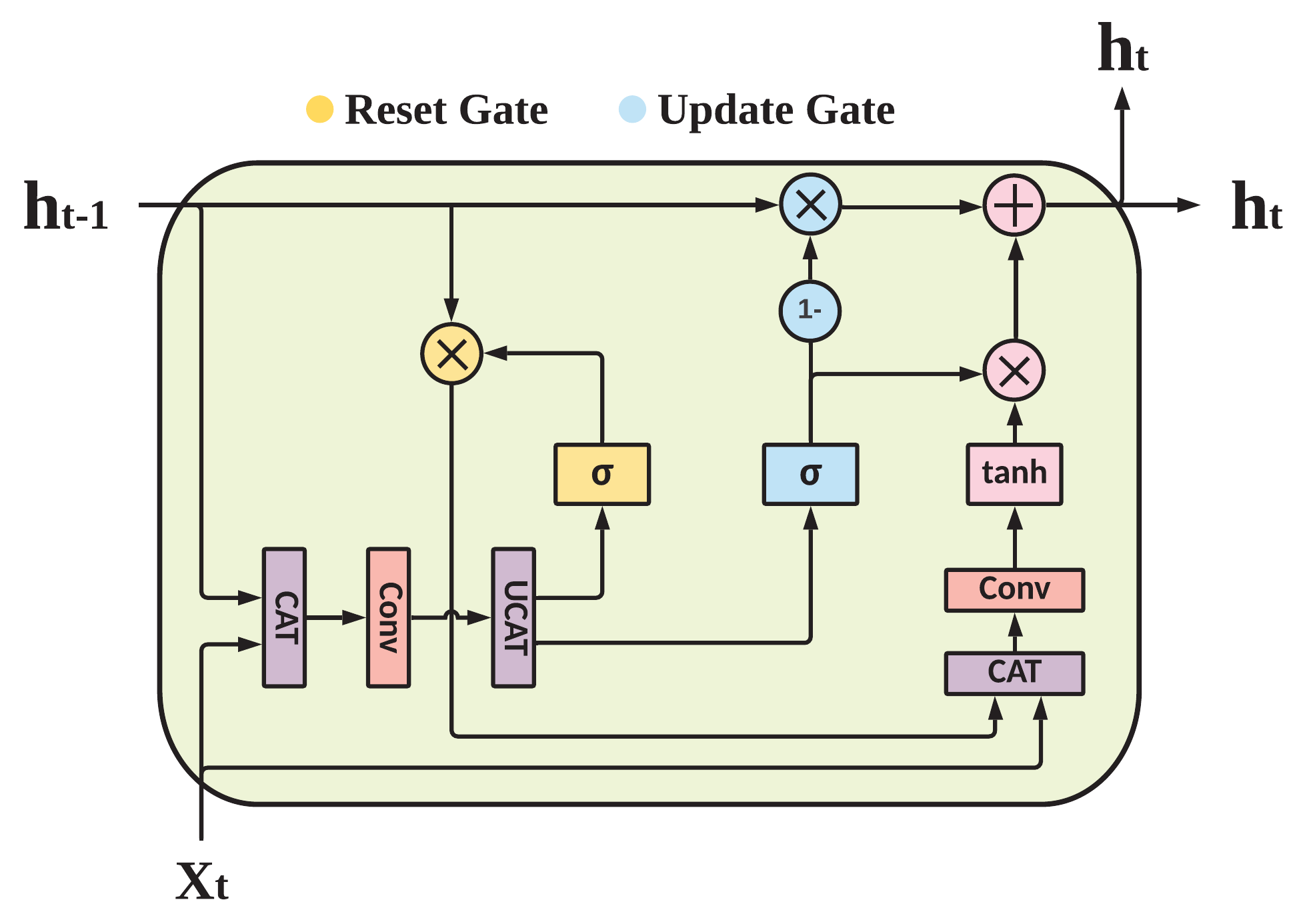}
  \caption{\small{The ConvGRU cell mainly consists of a reset gate and an update gate, where Conv denotes a 1D-convolutional kernel to capture dependencies in the feature level. CAT and UCAT denote concatenation and splitting operations, respectively.}}
  \label{fig:ConvGRUcell}
  \vspace{-0.5cm}
\end{figure}

Different from a vanilla GRU cell, the original dot product operations are replaced with convolution operations at each gate. To get a clearer picture of ConvGRU, we present the implementation of a cell with information flow and operations in Fig. \ref{fig:ConvGRUcell}. The detailed information flow of ConvGRU is shown in below equations:
\begin{equation}
z_{t} =\sigma\left(W_{conv1,1} * [h_{t-1}, x_{t}]\right) 
\end{equation}
\begin{equation}
r_{t} =\sigma\left(W_{conv1,2} * [h_{t-1}, x_{t}]\right)    
\end{equation}
\begin{equation}
\hat{h}_{t} =tanh\left(W_{conv2} * [\left(r_{t} \odot h_{t-1}\right), x_{t}] \right)    
\end{equation}
\begin{equation}
h_{t} =\left(1-z_{t}\right) \odot h_{t-1}+z \odot \hat{h}_{t} 
\end{equation}
where  $x_{t}$ denotes the input of current step, $h_{t}$ and $h_{t-1}$ denote the hidden state of current and previous step, respectively, $z_{t}$ denotes the update gate, $r_{t}$ denotes the reset gate, and $\hat{h}_{t}$ denotes the candidate hidden state. Moreover, $W_{conv1,1}$ and $W_{conv1,2}$ denote the convolutional kernel for each gate, and $W_{conv2}$ denotes the convolutional kernel used in the computation of candidate hidden state. Besides, $\sigma$ and $tanh$ denotes the sigmoid and hyperbolic tangent activation functions, $*$ is the convolution operation, $[\cdot]$ is the concatenation operation, $\odot$ is the hadamard product, or element-wise multiplication. It is noted that in the implementation of equations (1-2), the input and the hidden state of previous step are concatenated, convoluted with kernels, and then split into two parts as the reset gate and update gate.

%\begin{equation}
%z_{t} =\sigma\left(W_{h z} * h_{t-1}+W_{x z} * x_{t}+b_{z}\right) 
%\end{equation}

%\begin{equation}
%r_{t} =\sigma\left(W_{h r} * h_{t-1}+W_{x r} * x_{t}+b_{r}\right)    
%\end{equation}

%\begin{equation}
%\hat{h}_{t} =tanh\left(W_{h} *\left(r_{t} \odot h_{t-1}\right)+W_{x} * x_{t}+b\right)     
%\end{equation}

%\begin{equation}
%h_{t} =\left(1-z_{t}\right) \odot h_{t-1}+z \odot \hat{h}_{t} 
%\end{equation}

%\ref{fig:ConvGRUcell}.
%\vspace{-0.3cm}
 
\subsection{Variational ConvGRU}

%\ref{fig:ConvGRUvae}.
%\vspace{-0.3cm}
\begin{figure}
  \centering
  \includegraphics[height=4.5cm]{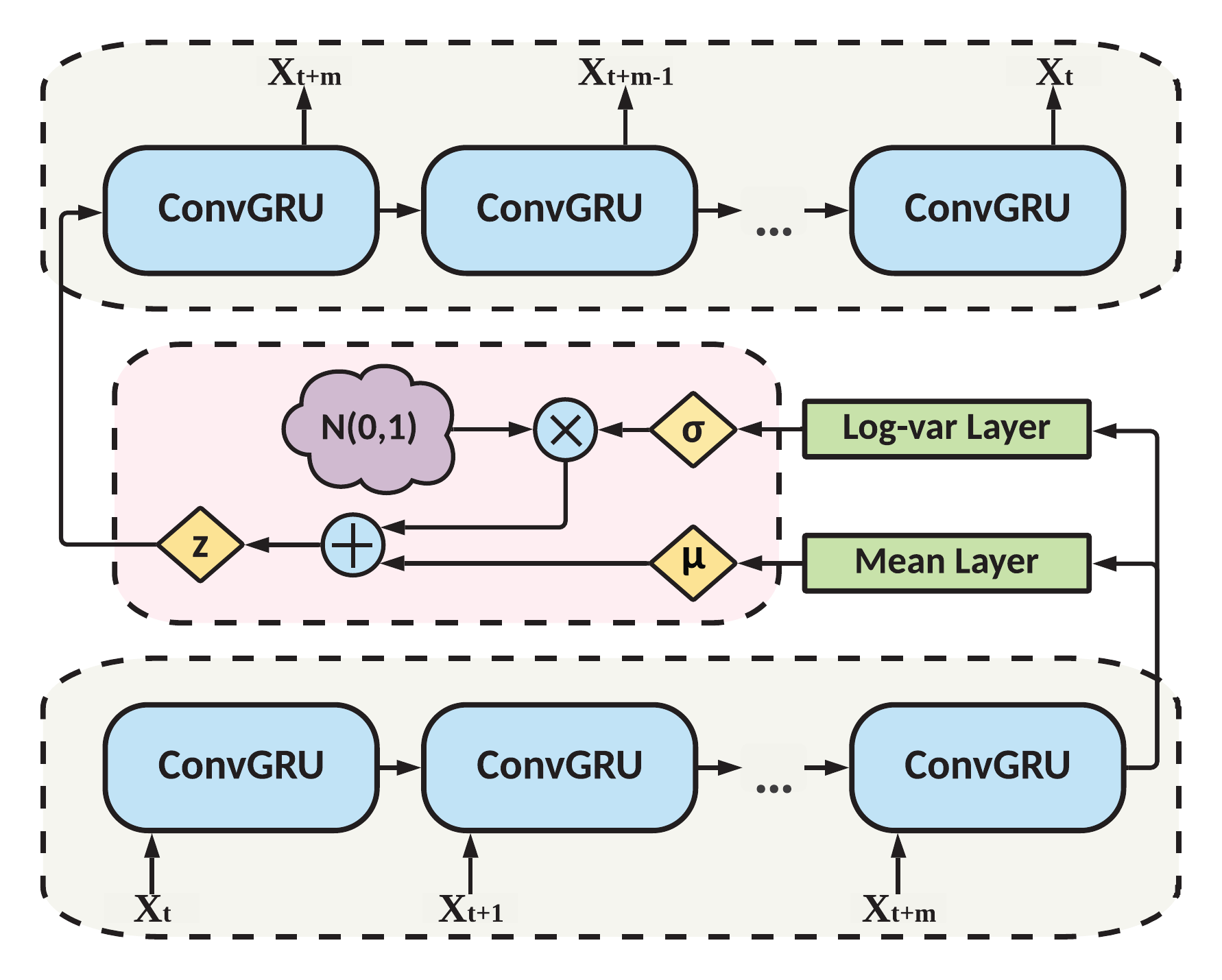}
  \caption{\small{The basic framework of ConvGRU-VAE.}}
  \label{fig:ConvGRUvae}
  \vspace{-0.5cm}
\end{figure}

VAE models the relationship between latent variable z and observed variable x. Specifically, the generative model is denoted as the joint distribution: $p_\theta$(x,z) = $p_\theta$(x$\mid$z)$p_\theta$(z), where $p_\theta$(x$\mid$z) is the likelihood of the data x given the latent z, which is distributed normally with multivariate unit Gaussian $\mathcal{N}(0,{\mathrm I})$. Since the intractable inference of the true posterior $p$(z$\mid$x), VAE uses the approximation posterior $q_\phi$(z$\mid$x) that can be assumed to be $\mathcal{N}(\mu_\phi, \sigma^2_\phi)$. The objective function is the variational lower bound of the marginal likelihood of data:
\begin{equation}
\begin{split}
    \log  p_\theta & (\mathrm x)  \geq \log p_\theta(\mathrm x) - {\rm KL} [q_\phi ({\mathrm z} | {\mathrm x}) \| p_\theta ({\mathrm z} | {\mathrm x})] \\ 
    & = \mathcal{L}({\theta, \phi; \mathrm x}) \\
    &= E_{z \sim q_{\phi}({\mathrm z} | \mathrm{x})} [\log p_{\theta}({\mathrm x} | \mathrm{z})] - {\rm KL} [q_\phi ({\mathrm z} | {\mathrm x}) \| p_\theta ({\mathrm z})]
\end{split}
\end{equation}
where $\mathcal{L}({\theta, \phi; \mathrm x})$ is the evidence lower bound (ELBO). To obtain the best parameter estimation, we can maximize the ELBO by means of Monta Carlo integration:
\begin{equation}
    E_{z \sim q_{\phi}({\mathrm z} | \mathrm{x})} [\log p_{\theta}({\mathrm x} | \mathrm{z})] \approx \frac{1}{L} \sum_{l=1}^L \log p_{\theta}({\mathrm x} | \mathrm{z}_{l})
\end{equation}

As for the second term (KL divergence), it could be explicitly equal to $\frac{1}{2} \sum_{j=1}^{J} (\mu_{j}^2 + \sigma_{j}^2 - log(\sigma_{j}^2) - 1)$ \cite{kingma2013auto}.

Conventional VAE is not a sequential model because it is merely composed of multilayer perceptrons. To deal with time sequences using VAE, we glue ConvGRU and VAE as illustrated in Fig. \ref{fig:ConvGRUvae}. We denote the logvar layer and mean layer in the figure as $f_{\phi}^{\sigma}$ and $f_{\phi}^{\mu}$, respectively. In detail, at time $t$, hidden features $\boldsymbol{\mathrm{h}_t}$ in spatio-tempoal domains are extracted from the last ConvGRU layer based on input observation $\boldsymbol{\mathrm{x}_t}$ and preceding hidden variable $\boldsymbol{\mathrm{h}_{t-1}}$. Then, Gaussian parameters of $\boldsymbol{\mathrm{x}_t}$ and $\boldsymbol{\mathrm{z}_t}$ are derived from the hidden features. Concretely speaking, we can obtain the means $\boldsymbol \mu_{\mathrm{x}_t}$ from a linear layer $f_{\phi}^{\mu}$. The reparameterization trick is adopted to generate a non-negative small number $\boldsymbol{\epsilon_{\mathrm{x}_t}}$, and we can get the standard deviations afterwards: $\boldsymbol{\sigma'_{\mathrm{x}_t}} = \boldsymbol{\sigma_{\mathrm{x}_t}} + \boldsymbol{\epsilon_{\mathrm{x}_t}}$, where $\boldsymbol{\sigma_{\mathrm{x}_t}} = f_{\phi}^{\sigma} (\boldsymbol{\mathrm{x}}_t)$. Now we have the latent variable $\boldsymbol{\mathrm{z}}_t \sim \mathcal{N}(\boldsymbol{\mu_{\mathrm{x}_t}, \sigma_{\mathrm{x}_t}'^2})$. The reconstructed sequence $\boldsymbol{\mathrm{x}_{t}'}$ is created on sampled $\boldsymbol{z}_t$ by a linear layer. Note that the output sequence of the ConvGRU decoder is in reverse order compared with the input sequence, which makes the optimization easier because the model can get off the ground by looking at low range correlations \cite{srivastava2015unsupervised}. Furthermore, as described in the problem statement, we focus more on the anomaly detection in the last timestamp of the input sequence. Therefore, only the hidden state of the last ConvGRU cell is employed in the decoder.

\section{Experiments} \label{sec:experiments}

In this section, we first present the implementation details, including data sources, model settings, and evaluation metrics. Then, we compare the performance between our models (non-federated and federated fashion) and baseline models. Furthermore, we conduct an extensive study on anomaly diagnosis under different setups.
\vspace{-0.15cm}

\subsection{Datasets} \label{sec:datasets}
To verify the effectiveness of our model, we conduct experiments on three datasets, including SMAP (Soil Moisture Active Passive satellite), MSL (Mars Science Laboratory rover), and Secure Water Treatment (SWaT) datasets. SMAP and MSL datasets come from telemetry data records of sensors in an individual spacecraft, where a set of associated telemetry values are collected, analyzed, and labeled by the experts in NASA, to identify anomalies that correspond to actual spacecraft issues \cite{hundman2018detecting}. SWaT dataset is collected from a real-world water treatment testbed at the Singapore University of Technology and Design, representing a scaled-down version of a realistic Cyber-Physical System whose behaviors can be learned and analyzed in order to assure safety and security in a network environment \cite{goh2016dataset}. As such, the training data is collected during two weeks of normal operation, and the testing data is collected in the following two days with a number of anomalous scenarios at different intervals. The multivariate time series in this dataset are presented based on one second and labeled as normal or abnormal at each time step. It is also noted that we down-sample the SWaT dataset by sequentially taking median values of each 10 instances for the purpose of efficient model training and evaluation. The statistical summary of three datasets is shown in Table \ref{Tlb: 2}.

\begin{table} 
  \setlength{\tabcolsep}{4.5pt}
  \small
  \caption{\small{Statistics of Datasets}}
  \label{Tlb: 2}
  \centering
  \begin{threeparttable}
     \begin{tabular}{lcccccl}
     \toprule
     \cmidrule{1-5}
     {Dataset} & {Train} & {Test} & {Features} & {Anomalies(\%)}\\

    \midrule
     SMAP & 135183 & 427617   & 25 & 13.13 \\
     MSL & 58317 & 73729  & 55  & 10.72 \\
     SWaT* & 49680 & 44992 & 51  & 11.98 \\
    \bottomrule
    \cmidrule{1-5}
    \end{tabular}
 \vspace{-0.12cm}
 
  \begin{tablenotes}
        \footnotesize
        \item[*] denotes the down-sampled version of original dataset.
      \end{tablenotes}
    \end{threeparttable}
 \vspace{-0.45cm}
\end{table}

\subsection{Experimental Setup}
In our experiments, we utilize a sequence-to-sequence learning method to capture the temporal representation at the training stage, while only the last step of the reconstructed sequence is used for performance evaluation and further studies at the testing stage. The pipeline for data preprocessing is presented as follows: Firstly, the original training data is split into 3:1 for the training set and validation set. Next, a min-max normalization is performed on the training set, validation set, and the given testing set, respectively. A single-step moving window with length of 10 steps as default, is then used to slice the training, validation, and testing sequences, after which they can be fed into the model. Because of the data heterogeneity presented in SMAP and MSL datasets (independent time windows, different window lengths, irregular intervals, various types of sensors), we apply an individual model to an individual channel as it is not possible to stack all channels' data in each dataset for training and testing.
 
\subsubsection{\textbf{Non-Federated Settings}} For model hyper-parameters, the kernel size of the convolutional gate in ConvGRU is 3. The hidden size of the ConvGRU cell is 128 as default. For training details, Averaged Stochastic Gradient Descent (ASGD) \cite{polyak1992acceleration} is used as the optimizer with learning rate 0.0001; the loss function consists of a reconstruction loss based on Mean Squared Error (MSE) and a Kullback–Leibler divergence loss measuring the difference between the latent and target distribution.

\subsubsection{\textbf{Federated Settings}} We use the default hyper-parameters as described above with optimizer SGD instead of ASGD. If not specified, the default local update epoch $E$ is set as $1$, and the number of clients $C$ is $3$. Each client exclusively owns data sampled from the training set in a non-iid manner \cite{mcmahan2017communication}. For SMAP and MSL, we first split the entire sequence in each channel into sub-sequences with a pre-defined window length and concatenate them, then divide the data into $S = W \times C$ shards of size $\lfloor \frac{N}{S} \rfloor$, and assign each client $W$ shards. $W$ is the hyperparameter that controls the size of data in each client and is equal to 2 by default, $C$ is the number of clients, and $N$ is the length of the entire data. Different from SMAP and MSL data, there is only one entity in the SWaT data that impedes conventional non-iid sampling based on skewed labels \cite{mcmahan2017communication}. Therefore, we let each client owns data of different time periods during the training stage on SWaT.
%the degree of data diversity among clients, 

\begin{table*} 
  \setlength{\tabcolsep}{7.0pt}
  \small
  \caption{\small{Performance comparison of different models}}
  \label{Tlb: 1}
  \centering
  \begin{threeparttable}
     \begin{tabular}{lccccccccl}
     \toprule
     \cmidrule{1-10}
     \multirow{2}{*}{Model} & \multicolumn{3}{c}{SMAP} & \multicolumn{3}{c}{MSL} & \multicolumn{3}{c}{SWaT}\\
    \cmidrule(lr){2-10}
     & Precision  & Recall  & F1  & Precision & Recall  & F1 & Precision & Recall & F1\\
    \midrule
     IF &  0.4423 & 0.5105 & 0.4671 &  0.5681 & 0.6740 & 0.5984 & 0.9620 & 0.7315 & 0.8311\\
     AE & 0.7216 & 0.9795 & 0.7776  & 0.8535 & 0.9748 & 0.8792  & \bf{0.9913} & 0.7040 & 0.8233\\
     
     LSTM-VAE & 0.7164 & 0.9875  & 0.7555  & 0.8599 & 0.9756  & 0.8537 & 0.7123 & 0.9258 & 0.8051\\
     DAGMM & 0.6334 & \bf{0.9984} & 0.7124 & 0.7562 & \bf{0.9803} & 0.8112 & 0.8292 & 0.7674 & 0.7971\\
     MAD-GAN &  0.8049 & 0.8214 & 0.8131 & 0.8517 & 0.8991 & 0.8747 & {0.9897} & 0.6374 & 0.7700 \\
     OmniAnomaly & 0.7416 & 0.9776 & 0.8434 & 0.8867 & 0.9117 & 0.8989 &0.7223 & \bf{0.9832} &0.8328\\
     USAD & 0.7697 & 0.9831 & 0.8186 &0.8810 & 0.9786 & 0.9109 &0.9870 &0.7402 &0.8460\\
    
    \midrule
     \textbf{ConvGRU-VAE} & \bf{0.9013} & {0.8320}  & \bf{0.8652}  & \bf{0.9211} & {0.9049}  &\bf{0.9129} & {0.9056} & {0.8632} & \bf{0.8839} \\
     \textbf{ConvGRU-VAE*} & {0.8906} & {0.9623}  & {0.9251}  & {0.9531} & {0.9815}  & {0.9671} & {0.8995} & {0.8743} & {0.8867}\\ 
     
    \midrule
     %\textbf{FedAnomaly} & \underline{0.9954} & {0.1491}  & {0.2594}  & \underline{0.9806} & {0.0130}  & {0.0257} & {0.9331} & {0.8164} & {0.8709} \\
     %\textbf{FedAnomaly} & {0.6739} & {0.4458}  & {0.5366}  & {0.5808} & {0.5857}  & {0.5833} & {0.9581} & {0.8127} & {0.8794} \\
     \textbf{FedAnomaly} & {0.6739} & {0.4458}  & {0.5366}  & {0.5808} & {0.5857}  & {0.5833} & {0.9331} & {0.8164} & {0.8709} \\
    \bottomrule
    \cmidrule{1-10}
    \end{tabular}
    
 \vspace{-0.1cm}
   \begin{tablenotes}
        \footnotesize
        \item[*] represents the model performance evaluated with a brute-force searched threshold based on the best F1 score.
        %\item[**] represents the model performance under federated mechanism.
      \end{tablenotes}
    \end{threeparttable}
 \vspace{-0.5cm}
\end{table*}

\subsection{Evaluation Metrics}
We use the common evaluation metrics in anomaly detection, which are Precision, Recall, and F1 score, to evaluate the performance of our models and baseline models over test data and ground truth labels:
\begin{equation}
\text{Precision} =\frac{\mathrm{TP}}{\mathrm{TP}+\mathrm{FP}}
\end{equation}
\begin{equation}
\text{Recall} =\frac{\mathrm{TP}}{\mathrm{TP}+\mathrm{FN}}
\end{equation}
\begin{equation}
\mathrm{F1} = 2 \times \frac{\text { Precision } \times \text { Recall }}{\text { Precision }+\text { Recall }}
\end{equation}
where TP (True Positive) denotes the the number of anomalies that are correctly detected, TN (True Negative) denotes the number of normal samples that are correctly classified, FP (False Positive) denotes the number of normal samples that are falsely detected as anomalies, and FN (False Negative) denotes the number of anomalies that are misclassified as normal samples. In real-world anomaly detection tasks, a higher precision often gives rise to more efficient operation, as fewer false alarms reduce the cost of examination for an entire system, which is expensive in many scenarios. On the other hand, a higher recall is also crucial as it indicates anomalies are more likely to be captured by the detector, including ones that might lead to system crashes. To have a more comprehensive evaluation toward anomaly detection, we give more consideration to the F1 score and try to avoid compromising too much of either precision or recall. Besides, we utilize the point-adjust method to evaluate the performance of models based on a contiguous-segment detecting strategy that is widely adopted in  \cite{zhao2020multi,audibert2020usad,su2019robust,chen2021learning,xu2018unsupervised}.
%state-of-the-art methods

\subsection{Comparison with Baseline Models} \label{sec:comparison}

\subsubsection{\textbf{Non-Federated Scenario}} We compare our model with 6 baseline models used in multivariate time series anomaly detection tasks, including Isolation Forest (IF) \cite{liu2008isolation}, the basic Autoencoder (AE) and state-of-the-art models: LSTM-VAE \cite{park2018multimodal}, DAGMM \cite{zong2018deep}, MAD-GAN \cite{li2019mad}, OmniAnomaly \cite{su2019robust} and USAD \cite{audibert2020usad}. We also present the evaluation results where the threshold is brute-force searched in terms of the best F1 score, in order to show the maximum capacity of our model. 
%Moreover, we implement the federated learning-based counterparts to evaluate the effectiveness of the federated mechanism. 
The performance of the detection models in terms of precision, recall, and F1 score are shown in Table \ref{Tlb: 1}. The results illustrate that our model generally outperforms the baseline models by achieving the highest precision score on the SMAP and MSL datasets, the third-highest recall score on SWaT dataset, and most importantly, the highest F1 score on the three datasets. It can also be observed that the false alarms and missing anomalies of our model are more comparable in comparison with those of other models, showing a better balance without compromising overall performance. 

The baseline methods consist of the conventional learning model and deep unsupervised learning models including generative models. Isolation Forest as a non-parametric method, uses a hyper-plane based on random features to split data space, which is not suitable for high-dimensional multivariate data and not sensitive to local anomalies. AE utilizes an encoding-decoding framework to reconstruct time series, based on which USAD model leverages the reconstruction-prediction and adversarial training to generate better representation for the downstream task. However, these methods reconstruct time series in a point-wise manner without capturing temporal dependencies, which limits the model capability and detection performance. DAGMM as a generative model, performs dimension reduction and density estimation of latent space in a joint manner for better representation learning and achieves the highest recall scores on SMAP and MSL datasets, but still fails to address temporal correlations, which gives rise to inferior overall performance. The other generative models based on recurrent neural networks gain advantages in handling temporal information and leverages either variational learning (LSTM-VAE, OmniAnomaly) or Generative Adversarial Networks (MAD-GAN). Nevertheless, the limit of these methods lies in either not taking feature-level correlations into consideration or not addressing them in an explicit way. Differently, our model not only captures the temporal dependencies and feature-level dependencies in input-to-state and state-to-state transitions, but also handles them in a joint manner. This joint information manipulation appears to be effective in the dependency modeling of multivariate time series and reports generally satisfactory results in downstream anomaly detection tasks.

To investigate the maximum capability of our model, we use the same output of ConvGRU-VAE as the anomaly score, but with a threshold that is brute-force searched based on calculated performance that gives the best F1 score. The results illustrate that our model with searched threshold gives obviously better recall and F1 scores on SMAP and MSL data. It is also worth noting that our model with original threshold selection demonstrates certain optimality as it approaches the best model on SWaT dataset. Generally speaking, our detection framework shows the potential to have better performance with more sophisticated threshold selection rules.

\begin{figure*}
  \centering
  \includegraphics[scale=0.34]{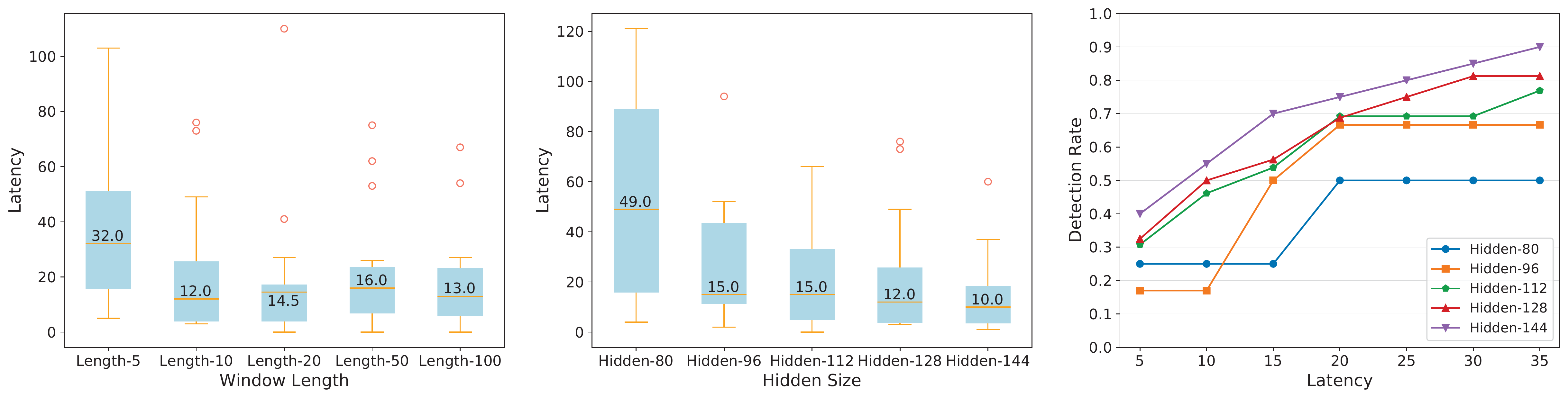}
  \caption{\small{The overview of latency distributions and the comparison under different latency constraints.}}
  \label{fig:latency}
  \vspace{-0.35cm}
\end{figure*}

\subsubsection{\textbf{Federated Scenario}} We adapt ConvGRU-VAE to the federated setting because of its superior which is demonstrated aforementioned. The last line in Tab. \ref{Tlb: 2} shows the best model performance (when C = 3) in federated scenarios. It is clear that FedAnomaly performs relatively worse on SMAP and MSL datasets than training on the entire data. %particularly in the metric of recall. Although it outperforms ConvGRU-VAE under non-federated scenario in terms of precision, it does no make sense in practice since the extremely low recall means the global model trained collaboratively can not detect anomalies effectively. 
Such an unsatisfactory result is caused by the extremely heterogeneous data distribution of the telemetry channels in the spacecraft \cite{hundman2018detecting}. In contrast, FedAnomaly has achieved similar performance of non-federated scenarios on SWaT dataset and surpasses most of baseline models regarding either precision and recall. 

Because of the clear description of labels and root-causes in the testing data of SWaT, we further explore the impact of different model parameters and federated settings in \cref{sec:non_federated_results} and \cref{sec:fedrated_results}, respectively. We also conduct latency analysis as described in \cite{ren2019time}, calculating the step difference between the first ground truth point and the first detected point in a true positive segment in the following text. 

\subsection{Exploration of Model Parameters and Latency Analysis} \label{sec:non_federated_results}

In this part, we further explore the effect of parameters in the single ConvGRU-VAE model and analyze the corresponding performance in terms of both detection results and detection latency. To be more specific, we perform a set of experiments on SWaT dataset where the anomalies are precisely labeled in a point-wise manner, based on different settings of window length and hidden size.

%\noindent \textbf{Exploration of window length:}
\subsubsection{\textbf{Exploration of window length}}
Firstly, we study how our model responds to different settings of window length. Table \ref{Tlb: 2} shows the detection performance based on different window lengths from $[5,10,20,50,100]$. It can be observed that our model adapts to different length of inputs and generally yields comparable results in terms of overall evaluation scores, where the best precision, best recall, and best F1 score is recorded at window length 5, 20 and 10 respectively. It can also be found that the precision and recall at window length 10 and 20, are relatively close without compromising overall F1 performance.

In real-world anomaly detection tasks, anomalies are more likely to occur in the form of contiguous segments than isolated points. Therefore, we also evaluate how our detection model triggers alerts under such consecutive scenarios. More specifically speaking, we involve the number of segments detected by our model and the corresponding detection latency in our discussion. The detection latency of each segment is calculated by the number of lagging steps between the first anomaly detected by our model and the first ground-truth anomaly in the segment. We can observe that our model detects relatively more segments with window length 10 and 20. Under each setting, we calculate the latency of all segments and present the box plots to demonstrate the overall latency distribution, as shown in the first subplot of figure \ref{fig:latency}. It is clear that our model performs better with a window length that is larger than 5, according to the median value and the distributed range of latency. This suggests that if the duration of the window is too short, each observation in the dataset could be less informative regarding quick detection of contiguous-segment anomalies. Considering all metrics discussed above, we conclude that our model performs better with window length 10 and 20 in terms of detection of overall anomalous patterns, detection of segments, and response speed.

\begin{table} 
  \setlength{\tabcolsep}{4.5pt}
  \small
  \caption{\small{Comparison of Window Length}}
  \label{Tlb: 2}
  \centering
     \begin{tabular}{lcccccl}
     \toprule
     \cmidrule{1-6}
      Window Length & 5 & 10 & 20 & 50 & 100 \\
    \midrule
     Precision & 0.9342 & 0.9057 & 0.8822 & 0.8555 & 0.9012 \\
     Recall & 0.8293 & 0.8632 &	0.8682 & 0.8487 & 0.8469\\
     F1 & 0.8786 &	0.8839 & 0.8751 & 0.8521 & 0.8732  \\
     Detected Segments & 13 & 16 & 18 & 15 & 13\\
    \bottomrule
    \cmidrule{1-6}
    \end{tabular}
 \vspace{-0.25cm}
\end{table}

%\noindent \textbf{Exploration of hidden size:}
\subsubsection{\textbf{Exploration of hidden size}}

\begin{table} 
  \setlength{\tabcolsep}{4.5pt}
  \small
  \caption{\small{Comparison of Hidden Size}}
  \label{Tlb: 3}
  \centering
     \begin{tabular}{lcccccl}
     \toprule
     \cmidrule{1-6}
      Hidden Size & 80 & 96 & 112  & 128 & 144 \\
    \midrule
     Precision & {0.9801} & 0.9829 & 0.8867 &	0.9057 & 0.8226 \\
     Recall & 0.6946 & 0.7246 &	0.8177 & 0.8632 & {0.8923} \\
     F1 & 0.8130 &	0.8342 & 0.8508 & {0.8839} &	0.8560   \\
     Detected Segments & 4 & 6 & 13 & 16 & {20}\\
    \bottomrule
    \cmidrule{1-6}
    \end{tabular}
 \vspace{-0.6cm}
\end{table}

\begin{table*}
\small
\caption{\small{Performance comparison of different federated scenarios}}
\centering
\begin{tabular}{lcccccccccccc}
\toprule
\multirow{2}{*}{Local Epoch} & \multicolumn{3}{c}{L = 1} & \multicolumn{3}{c}{L = 2} & \multicolumn{3}{c}{L = 3}  & \multicolumn{3}{c}{L = 5}  \\ 
\cmidrule{2-13} 
& P               & R               & F               & P                        & R               & F      & P      & R      & F      & P      & R      & F      \\ \midrule
C = 3                          & 0.8915          & 0.8418          & 0.8660          & \underline{0.9331}                   & 0.8164          & \bf{0.8709} & 0.7310 & \uuline{0.9054} & 0.8089 & 0.7606 & 0.8759 & 0.8142 \\
C = 5                          & \underline{0.9152}          & 0.8058          & 0.8570          & 0.9020                   & 0.8419 & \bf{0.8709} & 0.8057 & 0.8927 & 0.8469 & 0.7566 & \uuline{0.9043} & 0.8239 \\
C = 8                          & 0.8773          & 0.8275          & 0.8517          & \underline{0.9300}                   & 0.7970          & 0.8584 & 0.8475 & \uuline{0.9399} & 0.8476 & 0.9098 & 0.8297 & \bf{0.8679} \\
C = 10                         & \underline{0.9758} & 0.7858          & 0.8706 & 0.8378                   & 0.8490         & 0.8434 & 0.7718 & \uuline{0.8897} & 0.8266 & 0.9111 & 0.8527 & \bf{0.8809} \\
C = 20                         & 0.7874          & 0.8687 & 0.8260          & \underline{0.9407} & 0.8330          & \bf{0.8836} & 0.6989 & \uuline{0.9335} & 0.7994 & 0.8243 & 0.8623 & 0.8428 \\ 
\bottomrule
\label{tab:performance_fl}
\end{tabular}
 \vspace{-0.6cm}
\end{table*}

Besides window length, we also investigate the relationship between detection performance and model capacity measured by the hidden size. In general, a very small hidden dimension implies the insufficient model capacity to capture feature correlations and temporal dependencies, and often leads to inferior results. On the other hand, a very large hidden dimension introduces unnecessary redundancy which may impede effective representation learning and cause a degradation of model performance. To learn the impact of hidden size in our detection task, we perform the same set of evaluations as those in the above study. Table \ref{Tlb: 3} presents the detection performance based on different hidden size chosen from $[90, 96, 112, 128, 144]$. The results show that the model performance varies a lot in terms of detection scores. There is a clear trend of precision decrease and recall increase as the hidden size grows larger, while the best F1 performance is achieved at hidden size 128 with an obvious margin. This supports the above discussion and illustrates that a proper model capacity gains better robustness toward the balance of false alarms and missing anomalies.

Nevertheless, a larger hidden size within a certain range still gains favors in the segment detection. It is obvious that our model with hidden size 80 and 90 appears to be insufficient and performs poorly regarding the number of detected segments, while the model detects the most segments with hidden size 144. As for detection latency illustrated in the second subplot of figure \ref{fig:latency}, the insufficiency of hidden size shows an undesired distribution with worse median value and larger variance. Meanwhile, the latency of a model with a sufficient capacity generally distributes in a lower range and a more stable trend. To analyze the latency in an explicit manner, we also present the relationship between latency and detection rate, namely the percentage of detected anomalies with the constraint of different latency, as shown in the third subplot of Fig. \ref{fig:latency}. By and large, the model with a larger hidden size detects more anomalies when the tolerance of latency is small. Moreover, a larger hidden size consistently gives a better detection rate as latency constraints get looser. Therefore, the latency patterns shown and discussed above suggest the advantage of large model capacity in a certain range, which lies in the enhancement of faster and more accurate detection toward contiguous anomaly segments.

\subsection{Extensive Experiments on Federated Learning Mechanism} \label{sec:fedrated_results}

In this part, we fix the sequence length = 10, hidden size = 128, learning rate = 1e-5, ratio of clients $\gamma = 1$ and experiment with various scenarios determined by the local epoch $\textbf{L}$ and the number of clients $\textbf{C}$ in the federated mechanism. In detail, we let $\textbf{C} = \{3, 5, 8, 10, 20\}$ and $\textbf{L} = \{1, 2, 3, 5\}$.

\subsubsection{\textbf{Analysis of Performance}} Table \ref{tab:performance_fl} shows performance metrics under different federated scenarios. We use three types of highlights in the table: bold numbers, underlines, and double-underlines represent the best F1 scores, best precision, and best recall concerning each client setting, respectively. The best F1 scores and best precision scores are usually obtained together when L = 2 while the best recall scores are nearly attained when L = 3. Furthermore, the large local epoch often leads to better recall and worse precision no matter how many clients joining in collaborative training. For the scenarios with fewer clients (e.g. \textbf{C} $<$ 8), the precision declines gradually as the local epoch increases. However, no such intuitive pattern for the scenarios with much more clients. In other words, the performance can not be inferred given C and L, which may be caused by the distribution skew that then leads to weight divergence \cite{zhao2018federated}. In all scenarios, FedAnomaly achieves F1 scores of more than 0.8 that illustrates its robustness for time-series anomaly detection. 

Next, we present the detection latency results of FedAnomaly in Table \ref{tab:latency_fl}. There are two-digit tuples in the table where the first term is the number of adjusted segments while the second term is the average latency. We highlight the most detected segments and the lowest average latency for \textbf{C}. Obviously, 3 is the best local epoch for most client settings. However, recall Table \ref{tab:performance_fl}, precision score is usually not good when L = 3 that would result in many more false alarms.

\begin{table}
\small
\caption{\small{Latency analysis under different federated scenarios}}
\centering
\begin{tabular}{lccccc}
\toprule
C     & 3          & 5          & 8          & 10         & 20         \\ \midrule
L=1 & 14/28.49 & 12/31.83 & \textbf{13}/31.30 & 12/29.99 & 18/13.22 \\
L=2 & 14/26.71 & 14/29.99 & 12/\underline{15.49} & 14/19.92 & 12/28.58 \\
L=3 & \textbf{22}/\underline{12.86}& \textbf{21}/\underline{13.52} & 8/37.99  & \textbf{20}/\underline{14.99} & \textbf{25}/\underline{10.76} \\
L=5 & 19/13.52 & \textbf{21}/15.09 & \textbf{13}/32.92 & 15/28.93 & 17/16.94 \\ \bottomrule
\label{tab:latency_fl}
\end{tabular}
 \vspace{-0.75cm}
\end{table}

\begin{figure}[htbp]
    \centering
    \includegraphics[scale = .38]{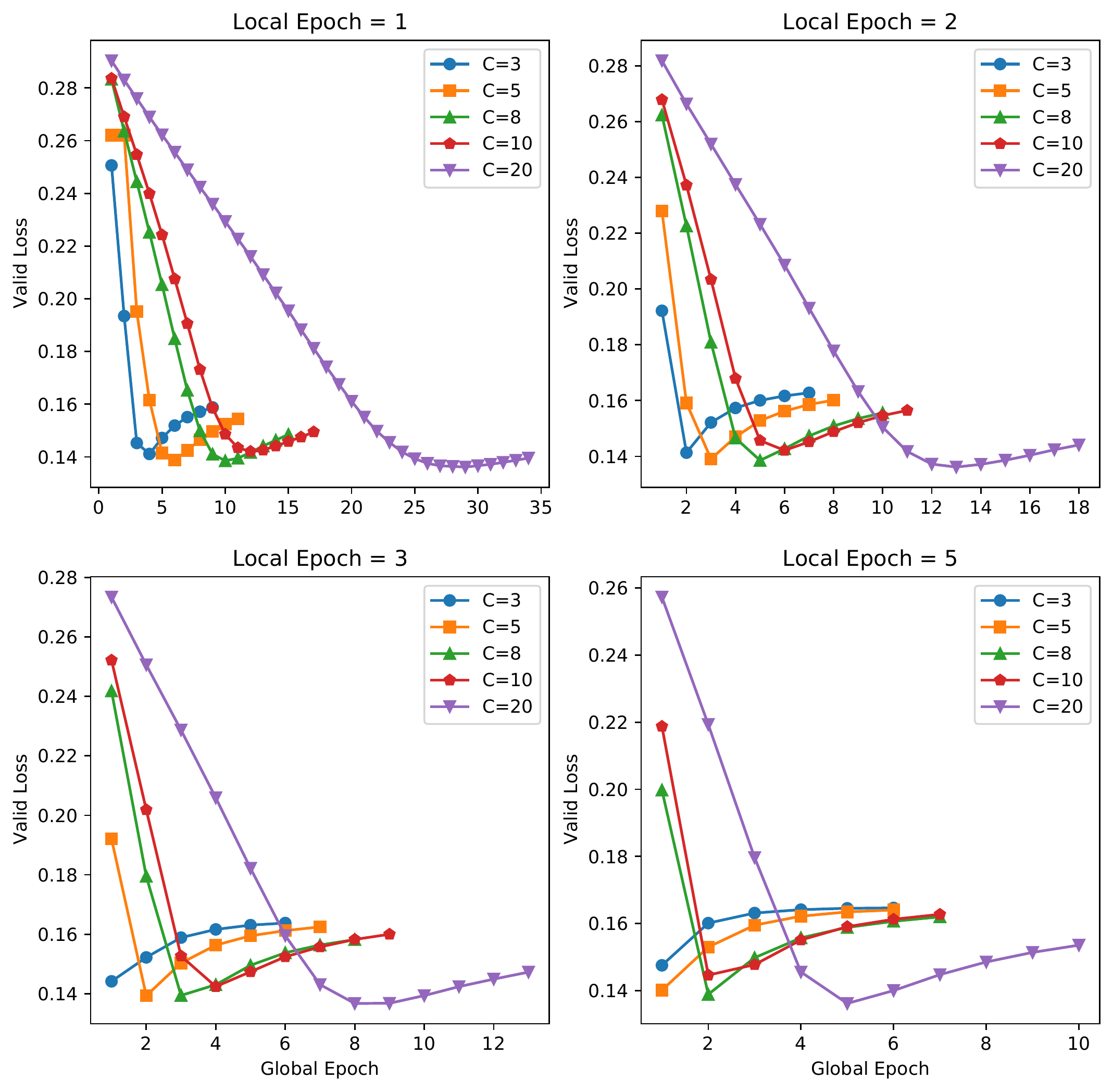}
    \caption{\small{Valid loss vs. global epoch (communication round).} }
    \label{fig:convergence}
     \vspace{-0.65cm}
\end{figure}

\subsubsection{\textbf{Analysis of Learning Curve}} We then explore the characteristics of the learning curve of FedAnomaly. We adopt the early stopping strategy in which FedAnomaly stops training if the reconstruction error (valid loss) on validation data on the cloud can not achieve the lowest value obtained in the previous communication rounds within the succeeding five rounds. Fig. \ref{fig:convergence} shows the validation curve in the training process. It illustrates that FedAnomaly needs more communication rounds to reach the lowest valid loss with the increase of the number of clients. From the perspective of the local epoch, the large value speeds up the convergence of the model particularly in the scenarios with more clients. For example, there are 25 communication rounds for obtaining the best model parameters when C = 20, L = 1, but only 5 rounds are needed if L = 5. Such a method is good for improving communication efficiency when a large number of edges collaboratively train a global model for time series anomaly detection. It is worth noting that in all settings, our model gets the similar best valid loss, approximately 0.14, that reflects both high capacity and robustness of our framework.

%\noindent \textbf{Analysis of Local Epochs:}

%\noindent \textbf{Analysis of Input Length:}

%\noindent \textbf{Analysis of Statistical Heterogeneity:} To answer the question that to what degree of data skew the FedAnomaly can still operate as normal, we perform multiple experiments with heterogeneous data distribution. 

\section{Conclusions and Future work} \label{sec:conclusion}

To address growing challenges within entity-level anomaly detection that stand to benefit significantly from unsupervised learning, we propose ConvGRU-VAE and its federated framework FedAnomaly, which work well for outlier detection tasks and outperform state-of-the-art approaches on three real-world multivariate time-series datasets. Through extensive experiments, we further explore the capacity of our detection framework and discuss the impact of different hyperparameter selection strategies. In the future, we plan to investigate solutions to tackle the data heterogeneity problem in the federated deep generative model for time series anomaly detection. %We are also interested in discovering anomaly interpretation solutions to annotate root causes.

\section*{Acknowledgment}
Effort sponsored by the Air Force under PIA FA8750-19-3-1000. The U.S. Government is authorized to reproduce and distribute copies for Governmental purposes notwithstanding any copyright or other restrictive legends. The views and conclusions contained herein are those of the authors and should not be interpreted as necessarily representing the official policies or endorsements, either expressed or implied, of the Air Force or the U.S. Government. This research was partially supported by the Center for Advanced Transportation Mobility (CATM), USDOT Grant No. 69A3551747125.

\bibliographystyle{IEEEtran}
\bibliography{Ref.bib}

\begin{thebibliography}{10}
\providecommand{\url}[1]{#1}
\csname url@samestyle\endcsname
\providecommand{\newblock}{\relax}
\providecommand{\bibinfo}[2]{#2}
\providecommand{\BIBentrySTDinterwordspacing}{\spaceskip=0pt\relax}
\providecommand{\BIBentryALTinterwordstretchfactor}{4}
\providecommand{\BIBentryALTinterwordspacing}{\spaceskip=\fontdimen2\font plus
\BIBentryALTinterwordstretchfactor\fontdimen3\font minus
  \fontdimen4\font\relax}
\providecommand{\BIBforeignlanguage}[2]{{%
\expandafter\ifx\csname l@#1\endcsname\relax
\typeout{** WARNING: IEEEtran.bst: No hyphenation pattern has been}%
\typeout{** loaded for the language `#1'. Using the pattern for}%
\typeout{** the default language instead.}%
\else
\language=\csname l@#1\endcsname
\fi
#2}}
\providecommand{\BIBdecl}{\relax}
\BIBdecl

\bibitem{ren2019time}
H.~Ren, B.~Xu, Y.~Wang, C.~Yi, C.~Huang, X.~Kou, T.~Xing, M.~Yang, J.~Tong, and
  Q.~Zhang, ``Time-series anomaly detection service at microsoft,'' in
  \emph{Proceedings of the 25th ACM SIGKDD International Conference on
  Knowledge Discovery \& Data Mining}, 2019, pp. 3009--3017.

\bibitem{lin2020anomaly}
S.~Lin, R.~Clark, R.~Birke, S.~Sch{\"o}nborn, N.~Trigoni, and S.~Roberts,
  ``Anomaly detection for time series using vae-lstm hybrid model,'' in
  \emph{ICASSP 2020-2020 IEEE International Conference on Acoustics, Speech and
  Signal Processing (ICASSP)}.\hskip 1em plus 0.5em minus 0.4em\relax IEEE,
  2020, pp. 4322--4326.

\bibitem{su2019robust}
Y.~Su, Y.~Zhao, C.~Niu, R.~Liu, W.~Sun, and D.~Pei, ``Robust anomaly detection
  for multivariate time series through stochastic recurrent neural network,''
  in \emph{Proceedings of the 25th ACM SIGKDD International Conference on
  Knowledge Discovery \& Data Mining}, 2019, pp. 2828--2837.

\bibitem{kingma2013auto}
D.~P. Kingma and M.~Welling, ``Auto-encoding variational bayes,'' \emph{arXiv
  preprint arXiv:1312.6114}, 2013.

\bibitem{mcmahan2017communication}
B.~McMahan, E.~Moore, D.~Ramage, S.~Hampson, and B.~A. y~Arcas,
  ``Communication-efficient learning of deep networks from decentralized
  data,'' in \emph{Artificial Intelligence and Statistics}.\hskip 1em plus
  0.5em minus 0.4em\relax PMLR, 2017, pp. 1273--1282.

\bibitem{yang2019federated}
Q.~Yang, Y.~Liu, T.~Chen, and Y.~Tong, ``Federated machine learning: Concept
  and applications,'' \emph{ACM Transactions on Intelligent Systems and
  Technology (TIST)}, vol.~10, no.~2, pp. 1--19, 2019.

\bibitem{chung2015recurrent}
J.~Chung, K.~Kastner, L.~Dinh, K.~Goel, A.~C. Courville, and Y.~Bengio, ``A
  recurrent latent variable model for sequential data,'' \emph{Advances in
  Neural Information Processing Systems}, vol.~28, pp. 2980--2988, 2015.

\bibitem{chen2021learning}
Z.~Chen, D.~Chen, Z.~Yuan, X.~Cheng, and X.~Zhang, ``Learning graph structures
  with transformer for multivariate time series anomaly detection in iot,''
  \emph{arXiv preprint arXiv:2104.03466}, 2021.

\bibitem{deng2021graph}
A.~Deng and B.~Hooi, ``Graph neural network-based anomaly detection in
  multivariate time series,'' in \emph{Proceedings of the AAAI Conference on
  Artificial Intelligence}, vol.~35, no.~5, 2021, pp. 4027--4035.

\bibitem{hundman2018detecting}
K.~Hundman, V.~Constantinou, C.~Laporte, I.~Colwell, and T.~Soderstrom,
  ``Detecting spacecraft anomalies using lstms and nonparametric dynamic
  thresholding,'' in \emph{Proceedings of the 24th ACM SIGKDD international
  conference on knowledge discovery \& data mining}, 2018, pp. 387--395.

\bibitem{hochreiter1997long}
S.~Hochreiter and J.~Schmidhuber, ``Long short-term memory,'' \emph{Neural
  computation}, vol.~9, no.~8, pp. 1735--1780, 1997.

\bibitem{chung2014gru}
J.~Chung, C.~Gulcehre, K.~Cho, and Y.~Bengio, ``Empirical evaluation of gated
  recurrent neural networks on sequence modeling,'' in \emph{NIPS 2014 Workshop
  on Deep Learning, December 2014}, 2014.

\bibitem{zong2018deep}
B.~Zong, Q.~Song, M.~R. Min, W.~Cheng, C.~Lumezanu, D.~Cho, and H.~Chen, ``Deep
  autoencoding gaussian mixture model for unsupervised anomaly detection,'' in
  \emph{International Conference on Learning Representations}, 2018.

\bibitem{xu2018unsupervised}
H.~Xu, W.~Chen, N.~Zhao, Z.~Li, J.~Bu, Z.~Li, Y.~Liu, Y.~Zhao, D.~Pei, Y.~Feng
  \emph{et~al.}, ``Unsupervised anomaly detection via variational auto-encoder
  for seasonal kpis in web applications,'' in \emph{Proceedings of the 2018
  World Wide Web Conference}, 2018, pp. 187--196.

\bibitem{li2019mad}
D.~Li, D.~Chen, B.~Jin, L.~Shi, J.~Goh, and S.-K. Ng, ``Mad-gan: Multivariate
  anomaly detection for time series data with generative adversarial
  networks,'' in \emph{International Conference on Artificial Neural
  Networks}.\hskip 1em plus 0.5em minus 0.4em\relax Springer, 2019, pp.
  703--716.

\bibitem{audibert2020usad}
J.~Audibert, P.~Michiardi, F.~Guyard, S.~Marti, and M.~A. Zuluaga, ``Usad:
  Unsupervised anomaly detection on multivariate time series,'' in
  \emph{Proceedings of the 26th ACM SIGKDD International Conference on
  Knowledge Discovery \& Data Mining}, 2020, pp. 3395--3404.

\bibitem{zhao2020multi}
H.~Zhao, Y.~Wang, J.~Duan, C.~Huang, D.~Cao, Y.~Tong, B.~Xu, J.~Bai, J.~Tong,
  and Q.~Zhang, ``Multivariate time-series anomaly detection via graph
  attention network,'' in \emph{2020 IEEE International Conference on Data
  Mining (ICDM)}.\hskip 1em plus 0.5em minus 0.4em\relax IEEE, 2020, pp.
  841--850.

\bibitem{li2019convergence}
X.~Li, K.~Huang, W.~Yang, S.~Wang, and Z.~Zhang, ``On the convergence of fedavg
  on non-iid data,'' in \emph{International Conference on Learning
  Representations}, 2019.

\bibitem{nguyen2019diot}
T.~D. Nguyen, S.~Marchal, M.~Miettinen, H.~Fereidooni, N.~Asokan, and A.-R.
  Sadeghi, ``D{\"i}ot: A federated self-learning anomaly detection system for
  iot,'' in \emph{2019 IEEE 39th International Conference on Distributed
  Computing Systems (ICDCS)}.\hskip 1em plus 0.5em minus 0.4em\relax IEEE,
  2019, pp. 756--767.

\bibitem{mcmahan2021advances}
H.~B. McMahan \emph{et~al.}, ``Advances and open problems in federated
  learning,'' \emph{Foundations and Trends{\textregistered} in Machine
  Learning}, vol.~14, no.~1, 2021.

\bibitem{liu2020deep}
Y.~Liu, S.~Garg, J.~Nie, Y.~Zhang, Z.~Xiong, J.~Kang, and M.~S. Hossain, ``Deep
  anomaly detection for time-series data in industrial iot: A
  communication-efficient on-device federated learning approach,'' \emph{IEEE
  Internet of Things Journal}, 2020.

\bibitem{zhao2019multi}
Y.~Zhao, J.~Chen, D.~Wu, J.~Teng, and S.~Yu, ``Multi-task network anomaly
  detection using federated learning,'' in \emph{Proceedings of the tenth
  international symposium on information and communication technology}, 2019,
  pp. 273--279.

\bibitem{chen2019network}
Y.~Chen, J.~Zhang, and C.~K. Yeo, ``Network anomaly detection using federated
  deep autoencoding gaussian mixture model,'' in \emph{International Conference
  on Machine Learning for Networking}.\hskip 1em plus 0.5em minus 0.4em\relax
  Springer, 2019, pp. 1--14.

\bibitem{shi2015convolutional}
X.~Shi, Z.~Chen, H.~Wang, D.-Y. Yeung, W.-K. Wong, and W.-c. Woo,
  ``Convolutional lstm network: A machine learning approach for precipitation
  nowcasting,'' \emph{Advances in neural information processing systems},
  vol.~28, 2015.

\bibitem{siam2017convolutional}
M.~Siam, S.~Valipour, M.~Jagersand, and N.~Ray, ``Convolutional gated recurrent
  networks for video segmentation,'' in \emph{2017 IEEE international
  conference on image processing (ICIP)}.\hskip 1em plus 0.5em minus
  0.4em\relax IEEE, 2017, pp. 3090--3094.

\bibitem{srivastava2015unsupervised}
N.~Srivastava, E.~Mansimov, and R.~Salakhudinov, ``Unsupervised learning of
  video representations using lstms,'' in \emph{International conference on
  machine learning}.\hskip 1em plus 0.5em minus 0.4em\relax PMLR, 2015, pp.
  843--852.

\bibitem{goh2016dataset}
J.~Goh, S.~Adepu, K.~N. Junejo, and A.~Mathur, ``A dataset to support research
  in the design of secure water treatment systems,'' in \emph{International
  conference on critical information infrastructures security}.\hskip 1em plus
  0.5em minus 0.4em\relax Springer, 2016, pp. 88--99.

\bibitem{polyak1992acceleration}
B.~T. Polyak and A.~B. Juditsky, ``Acceleration of stochastic approximation by
  averaging,'' \emph{SIAM journal on control and optimization}, vol.~30, no.~4,
  pp. 838--855, 1992.

\bibitem{liu2008isolation}
F.~T. Liu, K.~M. Ting, and Z.-H. Zhou, ``Isolation forest,'' in \emph{2008
  eighth ieee international conference on data mining}.\hskip 1em plus 0.5em
  minus 0.4em\relax IEEE, 2008, pp. 413--422.

\bibitem{park2018multimodal}
D.~Park, Y.~Hoshi, and C.~C. Kemp, ``A multimodal anomaly detector for
  robot-assisted feeding using an lstm-based variational autoencoder,''
  \emph{IEEE Robotics and Automation Letters}, vol.~3, no.~3, pp. 1544--1551,
  2018.

\bibitem{zhao2018federated}
Y.~Zhao, M.~Li, L.~Lai, N.~Suda, D.~Civin, and V.~Chandra, ``Federated learning
  with non-iid data,'' \emph{arXiv preprint arXiv:1806.00582}, 2018.

\end{thebibliography}

\end{document}